\documentclass{article}

\usepackage[preprint]{neurips_2024}

\usepackage[utf8]{inputenc} 
\usepackage[T1]{fontenc}    
\usepackage{hyperref}       
\usepackage{url}            
\usepackage{booktabs}       
\usepackage{amsfonts}       
\usepackage{nicefrac}       
\usepackage{microtype}      
\usepackage{xcolor}         
\usepackage{wrapfig}
\usepackage{graphicx}
\usepackage{amsmath}
\usepackage{subcaption}
\usepackage{makecell}
\usepackage{algorithm}
\usepackage{algpseudocode}
\usepackage{multirow}
\usepackage{colortbl}
\usepackage{enumitem}
\usepackage{cleveref}
\usepackage{tabstackengine}

\usepackage{amsmath}
\usepackage{amsthm}

\newtheorem{lemma}{Lemma}
\newtheorem{theorem}{Theorem}
\numberwithin{lemma}{section} 
\numberwithin{theorem}{section} 

\numberwithin{corollary}{section} 

\numberwithin{assumption}{section}

\newcommand{\secref}[1]{\S\ref{#1}}

\definecolor{commentcolor}{RGB}{0, 128, 0} 

\title{Achieving Sparse Activation in Small Language Models}

\author{%
	Jifeng Song, Kai Huang, Xiangyu Yin, Boyuan Yang and Wei Gao \\
	University of Pittsburgh\\
	\texttt{\{jifengsong, k.huang, eric.yin, by.yang, weigao\}@pitt.edu} \\
}

\begin{document}

\maketitle

\begin{abstract}
  Sparse activation, which selectively activates only an input-dependent set of neurons in inference, is a useful technique to reduce the computing cost of Large Language Models (LLMs) without retraining or adaptation efforts. However, whether it can be applied to the recently emerging Small Language Models (SLMs) remains questionable, because SLMs are generally less over-parameterized than LLMs. In this paper, we aim to achieve sparse activation in SLMs. We first show that the existing sparse activation schemes in LLMs that build on neurons' output magnitudes cannot be applied to SLMs, and activating neurons based on their attribution scores is a better alternative. Further, we demonstrated and quantified the large errors of existing attribution metrics when being used for sparse activation, due to the interdependency among attribution scores of neurons across different layers. Based on these observations, we proposed a new attribution metric that can provably correct such errors and achieve precise sparse activation. Experiments over multiple popular SLMs and datasets show that our approach can achieve 80\% sparsification ratio with $<$5\% model accuracy loss, comparable to the sparse activation achieved in LLMs. The source code is available at: \url{https://github.com/pittisl/Sparse-Activation}.
\end{abstract}

\section{Introduction}
In the era dominated by Large Language Models (LLMs), the recent emergence of Small Language Models (SLMs) presents an intriguing shift. Typical SLMs [\citenum{abdin2024phi, thawakar2024mobillama, team2024gemma}] are much more lightweight (e.g., with $<$3B parameters) than the existing LLMs \citenum{touvron2023llama, almazrouei2023falcon} (e.g., with $>$65B parameters), but can achieve on-par accuracy in simpler or specific tasks. However, inference with SLMs on resource-constrained mobile and embedded devices could still be too computationally expensive.

\begin{wrapfigure}{r}{0.35\textwidth}  
	\centering
	\includegraphics[width=0.32\textwidth]{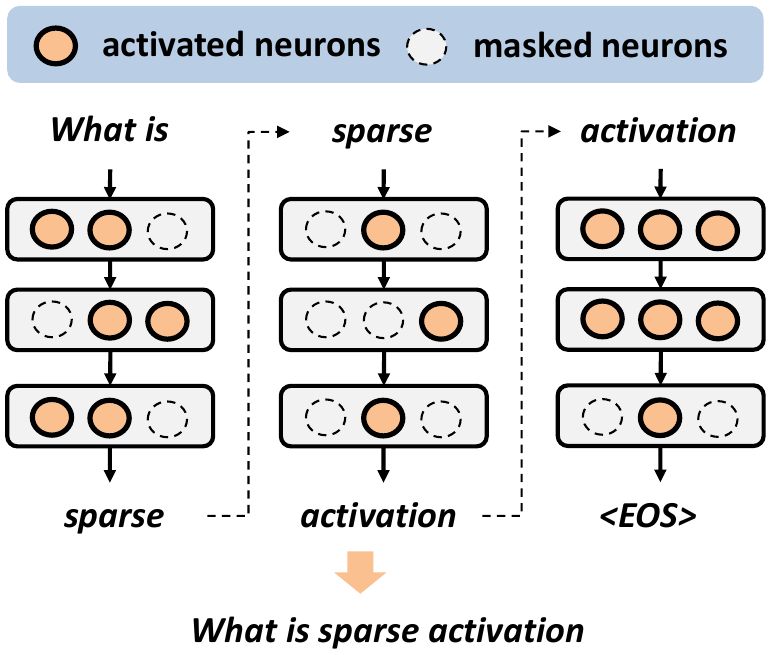}  
	\caption{Sparse activation for run-time improvement of inference performance}
	\vspace{-0.1in}
	\label{fig:sparse_activation}
\end{wrapfigure}

Research efforts have been made to reduce the inference cost of language models. Model compression techniques reduce the model size, via either pruning that sparsifies the model structures [\citenum{kurtic2024ziplm, ma2023llm, kurtic2024ziplm}], quantization that reduces the numerical precision of the model [\citenum{lin2023awq, kim2024memory, chee2024quip}], or knowledge distillation that migrates the general knowledge to a smaller model [\citenum{kang2024knowledge, zhao2023multistage}]. However, most of these methods require intensive model retraining, which is computationally expensive and cannot well adapt to different downstream tasks or input data [\citenum{michel2019sixteen, bansal2022rethinking}]. Other schemes of efficient attention algorithms [\citenum{dao2022flashattention}] and decoding techniques [\citenum{leviathan2023fast}] can also reduce the memory cost and computing latency in inference, but do not help reduce the model's redundancy. 
%

Sparse activation, as shown in Figure \ref{fig:sparse_activation}, can well complement the techniques above and enable run-time improvement of inference performance without any model retraining or adaptation efforts, by selectively activating only an input-dependent set of model's neurons that are most critical to inference with current input data\footnote{By re-implementation or using sparse computation APIs [\citenum{kurtz2020inducing}], the deactivated neurons can be masked to zero and further detached from the model's computing graph, to gain real wall-clock compute savings.}. Although sparse activation has been demonstrated to significantly reduce the memory cost and computing latency of inference in LLMs without impairing accuracy [\citenum{liu2023deja, song2023powerinfer}], whether it can be applied to SLMs still remains questionable. The major reason is that SLMs are much less over-parameterized than LLMs. Even neurons with small output magnitudes may still be important in inference, because they may largely affect the model's gradients across layers. Simply deactivating these neurons, as suggested by the existing work [\citenum{liu2023deja, song2023powerinfer}], may cause non-negligible accuracy loss. 
In this paper, we focus on achieving sparse activation in SLMs. Based on observations that deactivating neurons with small output magnitudes indeed results in large accuracy loss in SLMs, we experimentally showed that using gradient-based attribution scores to evaluate neurons' importance in inference and deactivate less important neurons is a more promising approach (\secref{subsec:attribution}), because such attribution scores precisely evaluate the impact of neuron deactivation on the model output. Some attribution schemes (e.g., Integrated Gradients [\citenum{sundararajan2017axiomatic,yvinec2022singe}]) integrate multiple gradients over interpolated input samples to ensure precise attribution scores, but also incur high computing costs. Other schemes provide more computationally efficient methods by calculating the attribution scores as the product of a neuron's gradient and output magnitude (Gradient $\times$ Output) [\citenum{liu2021group,lee2018snip}], which is the first-order approximation of the model output's change due to neuron deactivation. However, such attribution scores, when being applied to sparse activation, could result in large error due to the interdependency of neurons in one layer and across different layers. Such error may result in improper rankings of neuron's attribution scores and hence lead to suboptimal neuron activation (\secref{subsec:approximation_error}).

To effectively mitigate such attribution errors and achieve optimal sparse activation, we analytically quantified the lower and upper bounds of such error caused by inter-layer dependency among neurons in transformer-based SLM architectures, and further proposed to apply the expectation of such attribution error as a corrective term to the Gradient $\times$ Output (GxO) attribution metric (\secref{sec:method}). We evaluated the model performance when applying such corrected GxO metric for sparse activation, with multiple popular SLMs including Phi-1.5/2 [\citenum{gunasekar2023textbooks}] and MobiLlama-0.5B/1B [\citenum{thawakar2024mobillama}], and multiple question answering (QA) datasets including TruthfulQA [\citenum{lin2021truthfulqa}] and  YahooAnswersQA [\citenum{YahooAnswersQA}]. Our main findings are as follows:

\vspace{-0.08in}
\begin{itemize}
\item Our proposed attribution metric achieves high sparsity in SLMs, in both attention layers and MLP layers. It can deactivate up to 80\% of neurons in major SLM models while incurring $<$5\% model accuracy loss. Such sparisification ratio is similar to that reported in existing work for LLMs [\citenum{liu2023deja}], and allows significant memory savings and computing latency reduction.
\vspace{-0.17in}
\item We demonstrate good generality over different types of SLMs. On both models of the Phi and MobiLlama series, our proposed attribution metric outperforms baseline schemes in model accuracy by at least 25\%.
\vspace{-0.03in}
\item Our approach of applying the corrective term to neuron attribution metrics has high compute efficiency and incur a negligible amount of extra computing costs.
\end{itemize}
\vspace{-0.05in}
  
\vspace{-0.1in}
\section{Background and Motivation}

\subsection{SLMs with High Parameter Efficiency}
\vspace{-0.05in}
LLMs are known to be over-parameterized since their general knowledge is usually not fully useful to a specific downstream task (e.g., code generation [\citenum{gunasekar2023textbooks}]), and LLMs can hence be streamlined to achieve similar performance with significantly smaller sizes.  
To that end, existing SLM designs aim to achieve emergent abilities at a smaller scale with higher parameter efficiency. Some designs (e.g., Phi models [\citenum{phi-2}]) use better training strategies, such as involvement of synthetic datasets, to teach the model common-sense reasoning and general knowledge. Some other schemes (e.g., MobiLlama [\citenum{thawakar2024mobillama}]) adopt new architecture designs, such as parameter sharing between feed-forward networks (FFN) layers [\citenum{thawakar2024mobillama}] and parallel Multi-Head Attention (MHA) and FFN structures [\citenum{gunasekar2023textbooks, he2023simplifying}], to achieve better parameter efficiency without affecting model performance. Rotary Position Embeddings (RoPE) are also used in SLMs (e.g., Gemma [\citenum{team2024gemma}]), to enhance sequence modeling and improve the capture of positional information by incorporating rotational position encoding.


\vspace{-0.05in}
\subsection{Sparse Activation in Language Models} 
\vspace{-0.05in}
\label{subsec:attribution}
Sparse activation has been proved as effective in LLMs [\citenum{zhang2021moefication, liu2023deja}], particularly through the application of sparsity to FFN layers where ReLU activations often result in many neurons with zero output. Recent research efforts generalize the sparse activation based on neurons' output magnitudes [\citenum{zhang2024relu}] and deactivate neurons with small output magnitudes. To investigate the effectiveness such magnitude-based approach in SLMs, we examined the inference accuracy of different SLMs when we only activate neurons with the highest output magnitudes\footnote{In the rest of this paper, the terms ``neuron activation'' and ``neuron deactivation" are used interchangeably, as only activating a portion of neurons in the model is equivalent to deactivating other neurons in the model.}, using the TruthfulQA dataset [\citenum{lin2021truthfulqa}] and BLEU [\citenum{ke2023critiquellm}] as the accuracy metric. Since different SLMs may achieve variant levels of accuracy on this dataset, to ensure fair comparisons, for each SLM we first generate outputs from the fully activated model and then use these responses as the ground truth when evaluating the sparsely activated model. In this way, each fully activated model achieves nearly 100\% testing accuracy\footnote{With the TruthfulQA dataset, the model output will be mostly one or two words, and BLEU will be hence very low to penalize the short answer.}.

\begin{wrapfigure}{r}{0.35\textwidth}  
	\centering
	\includegraphics[width=\linewidth]{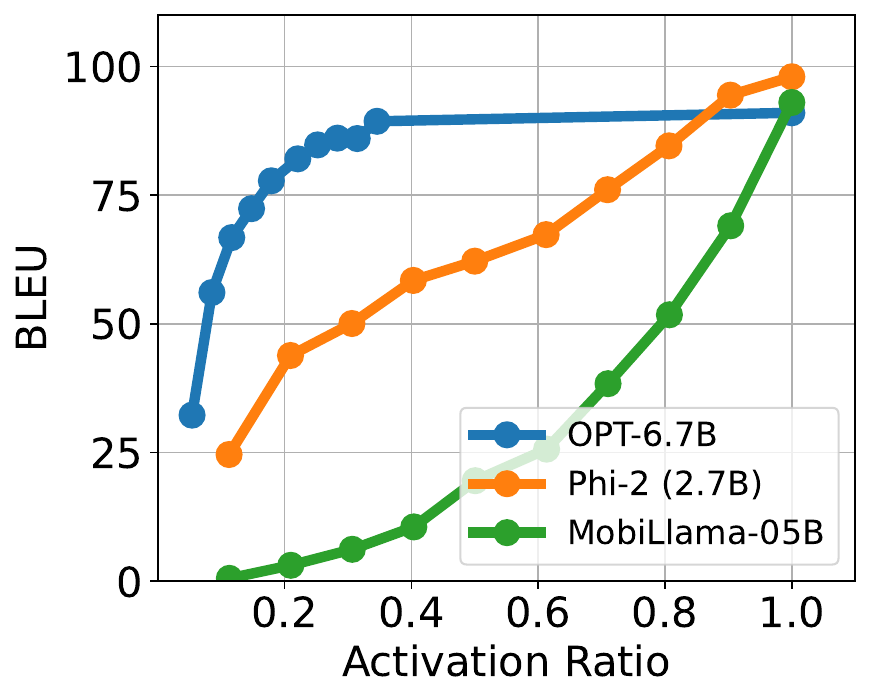}
	\caption{Accuracy-sparsity tradeoffs on different SLMs and LLMs with sparse activation based on neurons' output magnitudes}
	\label{fig: redundancy in SLMs and LLMs}
\end{wrapfigure}
As shown in Figure \ref{fig: redundancy in SLMs and LLMs}, OPT-6.7B is highly over-parameterized such that we only need to activate $<$40\% of neurons to achieve the maximum accuracy. In contrast, MobiLlama-0.5B and Phi-2 are much less over-parameterized, and both require almost all neurons to be activated to avoid accuracy loss. Even when a small percentage of neurons with the smallest magnitudes are deactivated, the model accuracy significantly drops. These results show that for SLMs, neurons' output magnitudes cannot precisely measure the neurons' importance in inference, and hence cannot be used as the metric for sparse activation. 

Instead, a better approach is to measure neurons' importance in inference with their attribution scores, and further use such attribution scores for sparse activation. In general, attribution methods quantify the correlation between input data, intermediate features and model output [\citenum{ivanovs2021perturbation}], and most recent methods calculate neurons' attribution scores from their gradients and outputs [\citenum{abhishek2022attribution}]. We investigated the effectiveness of representative gradient-based attribution metrics, as listed below, when evaluating a neuron's importance for sparse activation:

\begin{itemize}
	\item \textbf{Gradient $\times$ Output (GxO) [\citenum{liu2021group}]}: It calculates the first-order approximation of the change of model output when the neuron is deactivated, as $\partial F(x) / \partial x \cdot x$, where $x$ is the neuron's output scalar and $F$ is a function that maps neuron's output to the model output.
	\vspace{-0.05in}
	\item \textbf{SNIP [\citenum{lee2018snip}]}: It considers only the sensitivity of neuron's output change on the model output, and calculates the absolute value of GxO as $|\partial F(x) / \partial x \cdot x|$.
	\vspace{-0.05in}
	\item \textbf{Fisher information [\citenum{liu2021group}]}: It calculates the square value of SNIP as $|\partial F(x) / \partial x \cdot x|^2$, and hence ranks the importances of different neurons in the same ways as SNIP does.	
	\vspace{-0.05in}
	\item \textbf{Integrated Gradients (IG) [\citenum{sundararajan2017axiomatic}]}: It calculates the neuron's contribution to the change of model output by interpolating between $x$ and a baseline (usually zero output) and averaging the gradients at these interpolations, as $\frac{1}{n}\sum\nolimits_{k=1}^n \partial  F(\frac{k}{n} \cdot x)/\partial x \cdot x$.	
\end{itemize}

\begin{wrapfigure}{r}{0.35\textwidth}  
	\centering
		\vspace{-0.2in}
	\includegraphics[width=\linewidth]{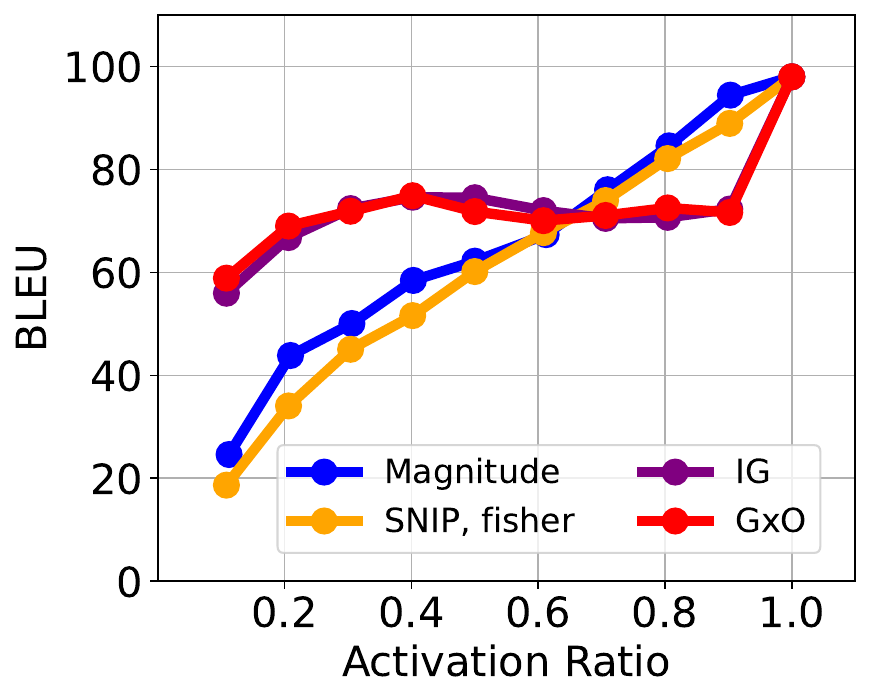}
		\vspace{-0.2in}
	\caption{Accuracy-sparsity tradeoffs using the Phi-2 model on the TruthfulQA dataset}
		\vspace{-0.1in}
	\label{fig: magnitude and the attribution method}
\end{wrapfigure}

Experiment results in Figure \ref{fig: magnitude and the attribution method} show that attribution-based sparse activation generally achieves higher model accuracy than the magnitude-based approach, with the same amount of neurons being activated. Among different attribution metrics, IG and GxO achieve the highest and very similar levels of model accuracy, because of taking both the magnitude and direction of gradients into the calculation of attributions. However, calculating IG is computationally expensive because of the large number of interpolations involved\footnote{At least 50 interpolations are suggested to ensure accuracy of calculating attributions [\citenum{sundararajan2017axiomatic}].}, and GxO is computationally efficient alternative that can be computed with a single forward and backward pass. In addition, since IG with a sufficient number of interpolations can by design precisely measure the neuron's attribution as the impact of neuron's deactivation on the model output [\citenum{sundararajan2017axiomatic}], such similarity in achieved model accuracy also verifies that GxO's first-order approximation to attribution is accurate, and the high-order reminders of this approximation is sufficiently small.

\subsection{Attribution Errors due to Interdependency}
\label{subsec:approximation_error}
\vspace{-0.05in}

\begin{wrapfigure}{r}{0.42\textwidth}  
	\centering
	\vspace{-0.1in}
	\includegraphics[width=0.42\textwidth]{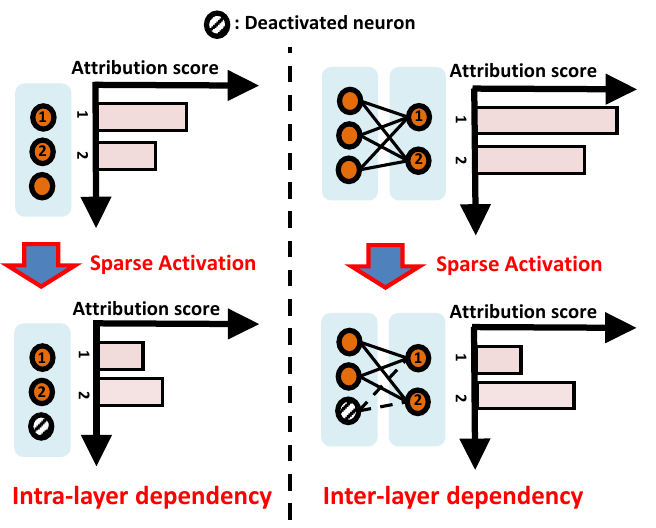}  
	\vspace{-0.1in}
	\caption{Interdependency of different neurons' attribution scores in sparse activation}
	\vspace{-0.1in}
	\label{fig: Two errors}
\end{wrapfigure}

When we calculate neurons' gradient-based attribution scores as described in Section \ref{subsec:attribution}, the attribution scores of different neurons are always interdependent. As shown in Figure \ref{fig: Two errors}, whenever some neurons are deactivated, such deactivation changes the attribution scores of other activated neurons, both in the same layer and in other subsequent layers. These changes, in many cases, could also change the rankings of neurons' attribution scores and hence result in suboptimal selection of neurons being deactivated, given a required activation ratio.

The main reason of such interdependency is the nonlinearity in SLMs. Let $F(x_1, x_2, ..., x_n)$ be the function that maps the neuron outputs $(x_1, x_2, ..., x_n)$ of a layer to the model output. Since $F$ is nonlinear, the gradient calculation of one neuron's output $x_1$, as $\partial F(x_1, x_2, ..., x_n) / \partial x_1$, always depends on other neurons' outputs, and hence deactivating any neuron will affect attribution scores of other neurons in the same layer. Similarly, deactivating neurons in one layer changes the neuron outputs and further attribution scores in subsequent layers.

\begin{figure}[ht]
	\hspace{-0.15in}
	\begin{minipage}[b]{0.35\linewidth}
		\centering
		\includegraphics[width=0.9\linewidth]{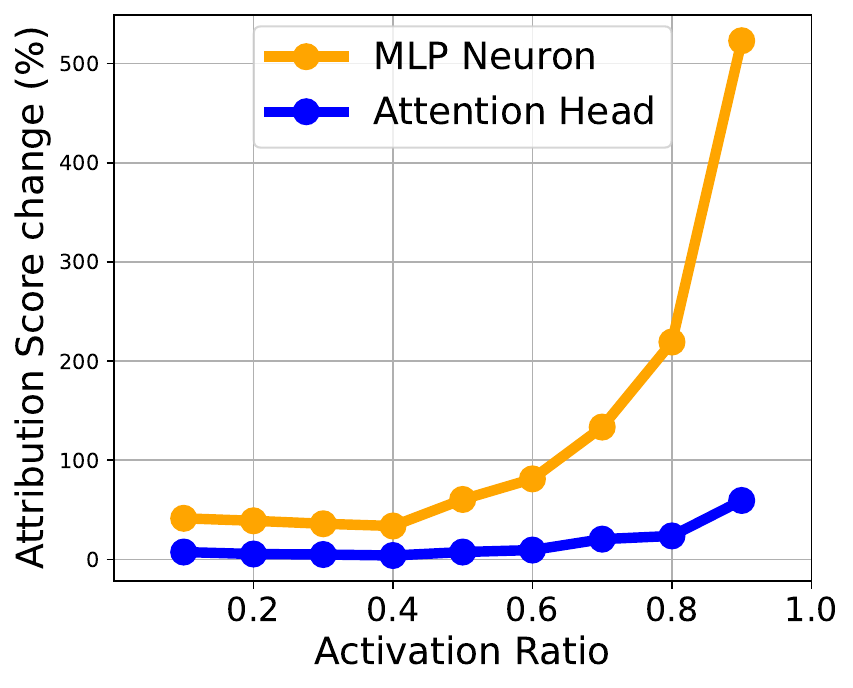}
		\vspace{-0.1in}
		\caption{The amount of neuron attribution scores changed by neuron deactivation}
		\vspace{-0.1in}
		\label{fig: Average_inter_layer_dependency_ratio}
	\end{minipage}	
	\hspace{0.1in}
	\begin{minipage}[b]{0.65\linewidth}	
		\vspace{-0.2in}		
		\includegraphics[width=\linewidth]{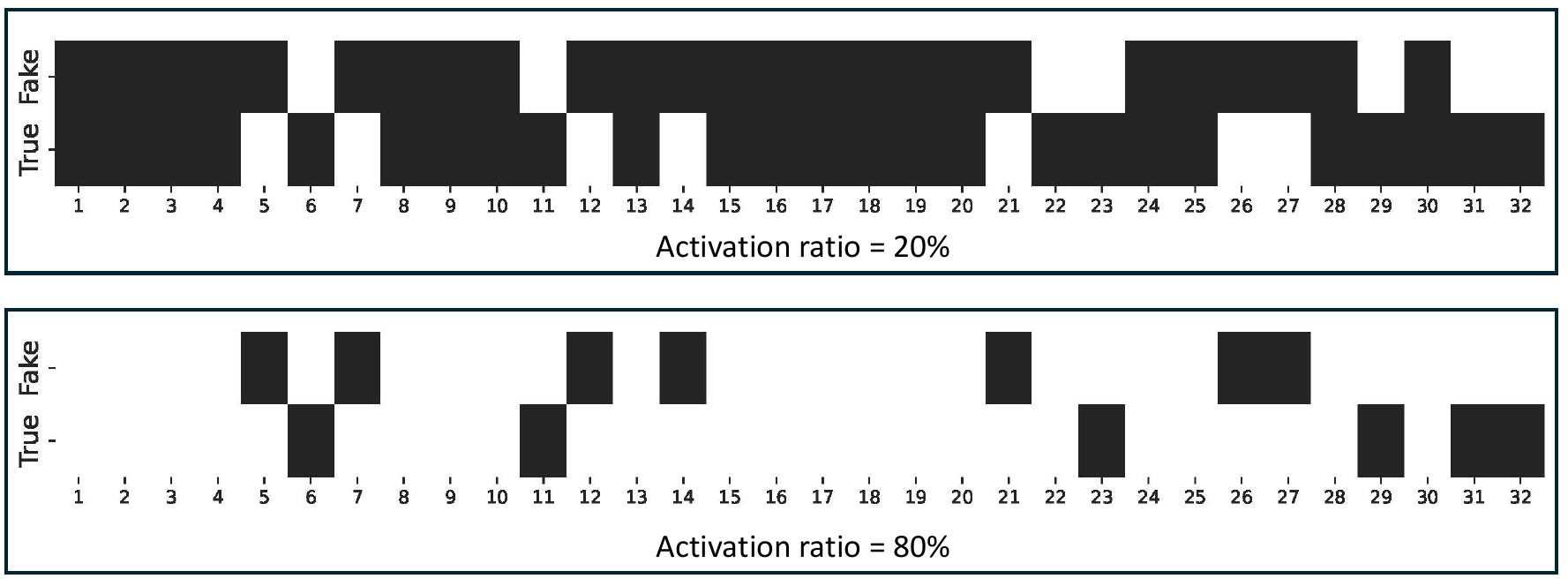}
		\vspace{-0.2in}
		\caption{Illustration of neuron deactivation among attention heads, using true (without attribution errors) and fake (with attribution errors) neuron attribution scores}
		\vspace{-0.1in}
		\label{fig: activation_ratio_illustration}
	\end{minipage}
\end{figure}

To explore the impact of such attribution errors, we apply different activation ratios on the Phi-2 model with the TruthfulQA dataset, and examine the amount of changes on the activated neurons' attribution scores due to deactivating other neurons. Results in Figure \ref{fig: Average_inter_layer_dependency_ratio} show that such impact significantly grows with higher activation ratios. The basic reason is that when the activation ratio is high, only few neurons are deactivated. Attribution errors, in this case, could easily change the ranking of those neurons with the lowest attribution scores and hence result in a different set of neurons being deactivated, as shown in Figure \ref{fig: activation_ratio_illustration}.  Meanwhile, we also found that attribution errors produce much higher impacts on MLP neurons, because the number of MLP neurons (e.g., 10,240 in Phi-2 model) is usually much larger than the number of attention heads (e.g., 32 in Phi-2 model), and the rank of MLP neurons' attention scores is hence easier to be changed.

To avoid such attribution errors, one intuitive approach is to individually calculate each neuron's attribution score and decide whether to deactivate this neuron, so that the calculation of any neuron's attribution is always based on the currently deactivated neurons. However, doing so is computationally expensive due to the large amount of neurons in SLMs and the difficulty of enforcing a specific activation ratio. Instead, we usually utilize vectorized computations supported in current deep learning APIs (e.g., TensorFlow and PyTorch) and calculate attributions of all neurons in one shot, but the aforementioned interdependency among neurons would largely affect the optimality of neuron deactivation.

We are then motivated to develop new techniques that can effectively mitigate these attribution errors and optimize the accuracy-sparsity tradeoffs in SLMs with proper sparse activation. In particular, the intra-layer dependency only reflects changes in the current layer's gradients because the outputs of neurons in the same layer are independent from each other. In contrast, the inter-layer dependency reflects changes in both the neuron outputs and gradients of the subsequent layer, as they all depend on the outputs of the previous layer. Hence, in the rest of this paper, we will mainly focus on mitigating the errors caused by inter-layer dependency.

\vspace{-0.1in}
\section{Mitigating the Attribution Errors in Sparse Activation}
\vspace{-0.1in}
\label{sec:method}
To mitigate the attribution errors, the most intuitive approach is follow the similar approach in model pruning [\citenum{yvinec2022singe}] and calculate neurons' attribution scores and apply sparse activation in a layer basis. More specifically, every time after sparse activation has been applied to one layer, we iteratively re-calculate the neurons' attribution scores in all the deeper layers. Doing so, however, still involve multiple forward and backward passes and incur large amounts of extra computing costs. Such amount of extra computing costs is also related to the model size and structure. For example, such cost on Phi-2 model with 32 layers is 1.3$\times$ more than that on Phi-1.5 model with 24 layers, and the cost on MobiLlama-1B model with 22 layers is another 1.8$\times$ higher due to its sequential transformer block architecture [\citenum{thawakar2024mobillama}].

Instead, our approach is to first analyze and quantify the attribution error caused by inter-layer dependency, and then mitigate such error by adding a corrective term onto each neuron's attribution calculated with the GxO metric, so that we can ensure proper sparse activation by calculating all neurons' attributions in one shot. More specifically, we formally proved the lower and upper bounds of the attribution error, and further provided practical methods of calculating and applying such corrective terms based on these bounds.

\subsection{Quantifying the Attribution Errors caused by Inter-Layer Dependency}

\begin{wrapfigure}{r}{0.42\linewidth}
	\vspace{-0.1in}
	\begin{minipage}[b]{0.7\linewidth}
		\centering
		{\fontsize{8}{9}\selectfont
			\begin{tabular}{>{\centering\arraybackslash}p{1.32cm}>{\centering\arraybackslash}p{4cm}}
				\toprule
				\textbf{Variables} & \textbf{Introduction}\\
				\midrule
				$x_i$ & Output of neuron $i$ in \(L_1\)\\
				\midrule
				$\mathbf{X}$ & Neuron output vector of \(L_1\)\\
				\midrule
				$\mathbf{\widetilde{X}}$ & Neuron output vector of \(L_1\) when $x_i$ is set as zero (deactivated)\\
				\midrule
				$g_j(\mathbf{X})$ & Output of neuron $j$ in \(L_2\) when the output of $L_1$ is $\mathbf{X}$ \\
				\midrule
				$g(\mathbf{X})$ & Neuron output vector of \(L_2\) when the output of $L_1$ is $\mathbf{X}$ \\
				\midrule
				$F(\cdot)$ & The function that maps the output of $L_1$ to the model output \\				
				\midrule
				$h(\cdot)$ & The function that maps the output of $L_2$ to the model output \\
				\midrule
				$S(F, x_i)$ & Attribution score of neuron $i$ in \(L_1\)\\
				\midrule
				$S(h, g_j(\mathbf{X}))$ & Attribution score of neuron $j$ in \(L_2\)\\	
				\midrule
				$S(F, \mathbf{X})$ & The sum of attribution scores of all neurons in $L_1$\\
				\midrule
				$S(h, g(\mathbf{X}))$ & The sum of attribution scores of all neurons in $L_2$\\
				\bottomrule
		\end{tabular}}	
	\vspace{-0.05in}	
		\centering\captionof{table}{List of notations} 
		\vspace{-0.1in}
		\label{tab: variables list}
	\end{minipage}
\end{wrapfigure}

Without loss of generality, we use a case of two layers in a SLM, namely $L_1$ and $L_2$, to quantify the attribution error caused by inter-layer dependency. $L_2$ is a deeper layer than $L_1$, and $L_2$'s neuron output hence depends on $L_1$'s neuron output $\mathbf{X}=(x_1, x_2, ..., x_{N_1})$ as $\mathbf{Y}=g(\mathbf{X})=(g_1(\mathbf{X}), g_2(\mathbf{X}), ..., g_{N_2}(\mathbf{X}))$, but $L_1$ and $L_2$ are not necessarily adjacent to each other. Hence, our analysis results and the proposed approach could generically applied to any neuron in the SLM. 


When we calculate neurons' attribution scores from the model's gradients and neurons' outputs as described in Section \ref{subsec:attribution}, attribution scores of neurons in $L_2$ will also depend on neuron outputs of $L_1$. We denote the attribution score of a neuron $x_i$ in $L_1$ as $S(F, x_i)$ and the attribution score of a neuron $x_j$ in $L_2$ as $S(h, g_j(\mathbf{X}))$, respectively, where $F(\mathbf{X})=h(\mathbf{Y})=h(g(\mathbf{X}))$ is the mapping to the SLM model output. Then, when we deactivate the neuron $i$ in $L_1$, the error of inter-layer dependency caused by this neuron deactivation in $L_1$ on the neurons' attribution scores in $L_2$ can be quantified as 
\begin{equation}
\sum_{j \in I(g(\mathbf{X}))} S(h, g_j(\widetilde{\mathbf{X}})) - \sum_{j \in I(g(\widetilde{\mathbf{X}}))} S(h, g_j(\widetilde{\mathbf{X}})),
\label{eq:quantification_error}
\end{equation}
where $\widetilde{\mathbf{X}}$ is identical to $\mathbf{X}$ except that $x_i$ is set as zero (deactivated), and $I(g(\mathbf{X}))$ denotes the set of neurons that have the smallest attributions in $L_2$ and are hence deactivated (with a given activation ratio), when the output vector of $L_1$ is $\mathbf{X}$. Note that since deactivating $x_i$ could change the ranking of neurons' attribution scores in $L_2$, usually $I(g(\mathbf{X}))\neq I(g(\widetilde{\mathbf{X}}))$. Eq. (\ref{eq:quantification_error}) hence represents the change of $L_2$'s deactivated neurons' cumulative attribution scores, when the neuron $i$ in $L_1$ is deactivated.

\vspace{-0.05in}
\subsection{The Corrective Term}
\vspace{-0.05in}
Intuitively, the quantification in Eq. (\ref{eq:quantification_error}) could allow us to calculate the corrective term that mitigates the attribution error caused by inter-layer dependency, but deactivating each neuron $i$ in $L_1$ will change $\widetilde{\mathbf{X}}$ and further require recalculation of $g(\widetilde{\mathbf{X}})$. Such requirement of recalculation prevents us from calculating the corrective terms for all neuron's attribution scores in one shot. Instead, we develop new methods that can enable such one-shot calculation of the corrective terms. To do so, we explore the lower and upper bounds of the quantified error in Eq. (\ref{eq:quantification_error}), and we first have the following lemma:

\vspace{0.05in}
\setcounter{lemma}{0}
\begin{lemma}
	\label{lemma:3.3}
	The error of inter-layer dependency caused by deactivating neuron $i$ in $L_1$, as quantified in Eq. (\ref{eq:quantification_error}), has a lower bound of 0, and an upper bound of $\left| S(F, \mathbf{X}) - S(F, \widetilde{\mathbf{X}}) \right|$, where $S(F, \mathbf{X}) = \frac{\partial{F}}{\partial{X}}\mathbf{X}^T = \sum\nolimits_{i=1}^{N_1} S(F, x_i)$ indicates the sum of all neurons' attribution scores in $L_1$. The proof is in Appendix \ref{appendix: C}.
\end{lemma}

In Lemma \ref{lemma:3.3}, the lower bound is reached when the ranking of neuron attribution scores in $L_2$ is not changed by $i$'s deactivation, corresponding to \(I(g(\mathbf{X})) = I(g(\widetilde{\mathbf{X}}))\) in Eq. (\ref{eq:quantification_error}), and the upper bound is reached when $I(g(\mathbf{X})) \cap I(g(\widetilde{\mathbf{X}}) = \emptyset$, i.e., the attribution error completely changes the set of neurons being deactivated in $L_2$. Hence, both the lower and upper bounds are tight. 

However, to practically compute the upper bound specified in Lemma \ref{lemma:3.3}, we still need to individually deactivate each neuron in $L_1$, in order to compute the corresponding $S(h, g(\widetilde{X}))$. In order to allow one-shot calculation of such bounds for all neurons, we further explore the correlation between the attribution scores of neurons in $L_1$ and $L_2$, and such correlation is described in the following lemma:

\vspace{0.05in}
\setcounter{lemma}{1}
\begin{lemma}
	\label{lem:3.1}
	The sum of attribution scores of neurons in $L_1$ is equal to that in $L_2$. That is, if we denote $S(F, \mathbf{X}) = \frac{\partial{F}}{\partial{X}}\mathbf{X}^T = \sum\nolimits_{i=1}^{N_1} S(F, x_i)$ and $S(h, g(\mathbf{X})) = \frac{\partial{h}}{\partial{g(\mathbf{X})}}\mathbf{g(\mathbf{X})}^T = \sum\nolimits_{i=1}^{N_2} S(h, g_i(x))$, we have
	\begin{equation}
	S(F, \mathbf{X})=S(h, g(\mathbf{X})).
	\end{equation}
	The proof is in Appendix \ref{appendix: A}
\end{lemma}

With Lemma \ref{lemma:3.3} and Lemma \ref{lem:3.1}, we can correlate the error of inter-layer dependency caused by neuron deactivation in $L_1$, as specified in Eq. (\ref{eq:quantification_error}), to the neurons' attribution scores in $L_1$. Then, the following theorem provides the bounds of such error that are decided by the neuron gradients in $L_1$:

\vspace{0.05in}
\setcounter{theorem}{2}
\begin{theorem}
	\label{Theorem:3.4}
	The error of inter-layer dependency caused by deactivating neuron $i$ in $L_1$ has a lower bound of 0, and a upper bound of ${|x_i|} \cdot \sqrt{\sum_{k=1}^{N_1}  (\frac{\partial F}{\partial x_k})^2}$, where $x_k$ is the output of another neuron $k$ in $L_1$. The proof is in Appendix \ref{appendix: D}
\end{theorem}

\begin{wrapfigure}{r}{2.2in}
	\centering
	\vspace{-0.4in}
	\includegraphics[width=\linewidth]{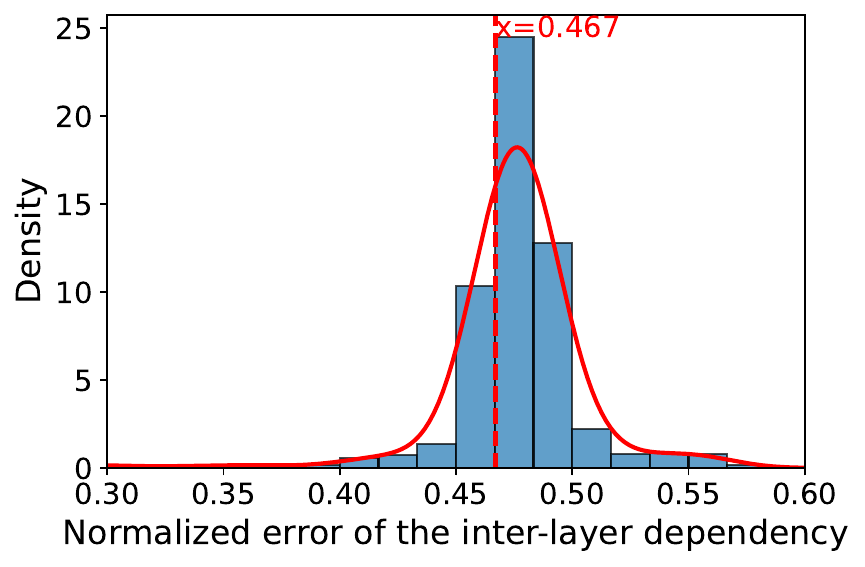}
	\vspace{-0.2in}
	\caption{The distribution of attribution error of inter-layer dependency}
	\vspace{-0.3in}
	\label{fig:importance_scores_change}
\end{wrapfigure} 

Being similar to Lemma \ref{lemma:3.3}, both bounds in Theorem \ref{Theorem:3.4} are also tight. 

Based on Theorem \ref{Theorem:3.4}, with the knowledge about the distribution of attribution errors overall all neurons in the model, we can calculate the corrective term being applied to each neuron's attribution score as the expectation of such distribution. According to our experiment results shown in Figure \ref{fig:importance_scores_change} with Phi-2 model and TruthfulQA dataset, such distribution can be well approximated by a truncated normal distribution with a $>$99\% confidence interval. Hence, we can calculate the corrective term as 
\begin{equation}
C(i)=\frac{1}{2} \cdot {|x_i|} \cdot \sqrt{\sum_{k=1}^{N_1}  (\frac{\partial F}{\partial x_k})^2}.
\label{eq:corrective_term}
\end{equation}

This corrective term is only related to the output magnitudes and gradients of neurons, and hence such corrective terms of all neurons can be computed in one shot with vectorized computations enabled in the existing deep learning APIs.

\vspace{-0.05in}
\subsection{Practical Operations}
\vspace{-0.05in}

\begin{wrapfigure}{r}{0.35\textwidth}  
	\centering
	\vspace{-0.3in}
	\includegraphics[width=\linewidth]{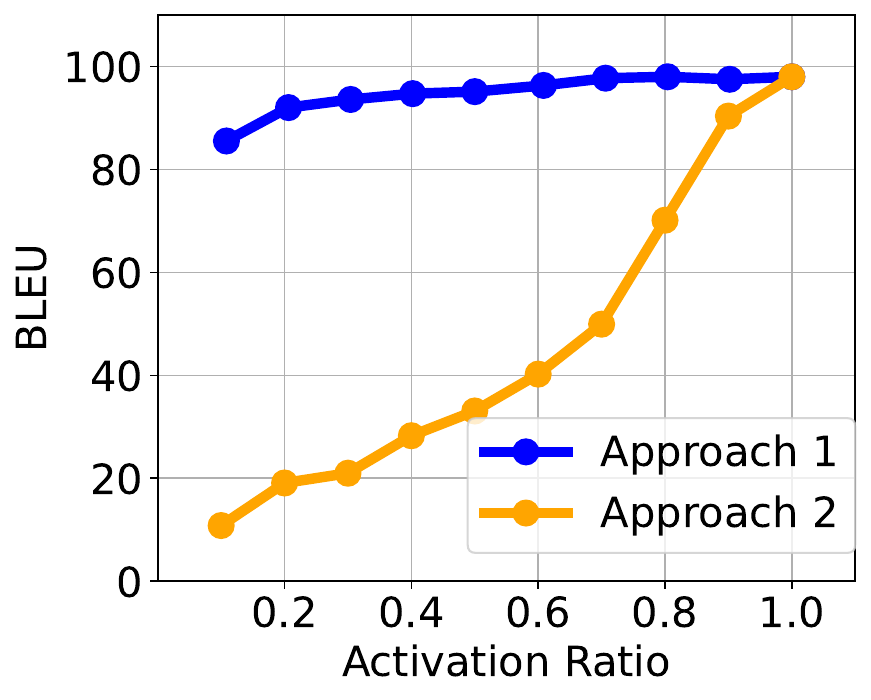}
	\vspace{-0.2in}
	\caption{Setting same activation ratio and threshold for each layer}
	\vspace{-0.1in}
	\label{fig: activation_ratio_threshold}
\end{wrapfigure}

In practical sparse activation, we can either activate the same percentage of neurons in each layer by applying a layer-specific threshold on neurons' attribution scores (Approach 1), or applying a uniform threshold on attribution scores in all layers and hence activate different percentages of neurons in different layers (Approach 2). To compare the actual model performance with these two schemes, we conducted preliminary experiments using Phi-2 model and the TruthfulQA dataset. Results in figure \ref{fig: activation_ratio_threshold} show that applying a layer-specific threshold to activate the same percentage of neurons in each layer always leads to better model performance with different activation ratios. The basic reason is that our corrective term is calculated from the neuron's gradients and output magnitudes, both of which could vary a lot across different layers, and the attribution scores of neurons in different layers are hence not comparable when such corrective term is applied to neurons' attribution scores. Instead, layer-wise decisions on neuron activation are more appropriate, and layer-wise neuron activation also allows easy enforcement of any specific activation ratio.

\vspace{-0.1in}
\section{Experiments}
\vspace{-0.1in}

\textbf{Models and datasets.} We evaluate the accuracy of multiple sparsely activated SLMs, when neuron activations are decided by applying the corrective term proposed in Eq. (\ref{eq:corrective_term}) to the GxO attribution metric [\citenum{liu2021group}]. Our experiments use the following SLMs:
\vspace{-0.05in}
\begin{itemize}
	\item  \textbf{Phi-1.5} [\citenum{gunasekar2023textbooks}] is a general-purpose SLM with 1.3B parameters developed by Microsoft for a large collection of applications, including text completion, translation, and sentiment analysis.
	\vspace{-0.15in}
	\item  \textbf{Phi-2} [\citenum{gunasekar2023textbooks}] is a newer version in the Phi series of SLMs, built upon the capabilities of Phi-1.5 with larger model sizes (2.7B parameters) and enhanced model accuracy.
	\vspace{-0.05in}
	\item  \textbf{MobiLlama-0.5B} [\citenum{thawakar2024mobillama}] is a SLM design that initiates from a larger model (LLaMA-7B) applies a careful parameter sharing scheme to reduce both the pre-training and deployment costs on mobile and low-power devices.		
	\vspace{-0.05in}
	\item  \textbf{MobiLlama-1B} [\citenum{thawakar2024mobillama}] is a larger version in the MobiLlama series of SLMs, offering superior accuracy and depth in language processing tasks.
\end{itemize}
\vspace{-0.05in}

Our experiments focus on examining the model accuracy on the question answering (QA) task in the natural language domain, which is one of the main targeted tasks of most SLM designs. More specifically, our experiments use the following two QA datasets:
\vspace{-0.05in}
\begin{itemize}
	\item \textbf{TruthfulQA} [\citenum{lin2021truthfulqa}] is a well-known QA dataset to measure whether a language model is truthful in generating answers to questions. It contains 817 QA pairs in 38 categories, including health, law, finance and politics.	
	\vspace{-0.05in}
	\item \textbf{YahooAnswersQA} [\citenum{YahooAnswersQA}] is a QA dataset that is derived from the Yahoo Answers platform, which was a popular community-driven QA website. It contains 87.4k QA pairs and covers topics of Science \& Mathematics, Society \& Culture, and Food \& Drink.	
\end{itemize}
\vspace{-0.05in}

\textbf{Baseline schemes.} We compare the accuracy of SLMs being sparsely activated by using our corrected GxO metric with that using other attribution metrics listed in Section \ref{subsec:attribution}, including Integrated Gradients (IG) [\citenum{sundararajan2017axiomatic}], SNIP [\citenum{lee2018snip}] and Fisher [\citenum{liu2021group}]. We also include two other naive baselines, i.e., directly using the neurons' magnitudes and gradients as the metric for sparse activation.

\textbf{Evaluation setup.} In our experiments, we measure the model accuracy using the BLEU (Bilingual Evaluation Understudy) score [\citenum{ke2023critiquellm}], which is a widely used metric to evaluate the quality of text by language models, as the similarity of the generated text to one or more reference texts. A higher BLEU score (in percentage) indicates better similarity between the generated text and the reference text. Since different models may exhibit highly variant levels of accuracy on the same dataset. In order to ensure fair comparisons across different models and datasets, instead of comparing the model's generated answers to the ground truths given by the dataset, we compare the sparsely activated model's generated answer with the answer generated by the fully activated model under the same setting. Our experiments are conducted using a cloud GPU server with one NVidia H100 80GB GPU. Our details about our evaluation and implementation setup are in Appendix \ref{appendix: F}.



\begin{table*}[htbp]
	{\fontsize{8}{10}\selectfont
		\hspace{-0.1in}
		\begin{tabular}{crrrrrrrr}
			\toprule
			\textbf{Metric} & \textbf{AR=10\%} & \textbf{AR=20\%} & \textbf{AR=30\%} & \textbf{AR=40\%} & \textbf{AR=50\%} & \textbf{AR=60\%} & \textbf{AR=80\%} & \textbf{AR=100\%}\\
			\midrule
			Gradient & 5.8 & 5.4 & 5.7 & 5.2 & 5.0 & 5.2 & 5.6 & 98.0 \\
			Magnitude & 24.6 & 43.9 & 50.1 & 58.4 & 62.2 & 67.3 & 84.6 & 98.0 \\
			SNIP/Fisher & 18.7 & 34.1 & 45.1 & 51.6 & 60.2 & 67.7 & 82.1 & 98.0 \\
			IG & 56.0 & 66.8 & 72.4 & 74.6 & 74.5 & 72.0 & 70.7 & 98.0 \\
			GxO & 58.9 & 69.0 & 71.9 & 74.9 & 71.8 & 70.0 & 72.5 & 98.0 \\			
			Corrected GxO & \textbf{\underline{85.5}} & \textbf{\underline{92.0}} & \textbf{\underline{93.6}} & \textbf{\underline{94.7}} & \textbf{\underline{95.1}} & \textbf{\underline{96.3}} & \textbf{\underline{98.0}} & 98.0 \\
			\bottomrule
	\end{tabular}}
	\caption{Accuracy of sparsely activated SLMs with different activation ratios (ARs). The Phi-2 model on the TruthfulQA dataset is used.} 
	\label{tab:basic_results}
\end{table*}

\vspace{-0.05in}
\subsection{Accuracy of Sparsely Activated SLMs}
\vspace{-0.05in}

We first evaluate the model accuracy with different activation ratios (ARs), using the Phi-2 model on the TruthfulQA dataset. Results in Table \ref{tab:basic_results} show that, when applying our proposed corrective term onto the GxO metric, our approach generally achieves much higher model accuracy than all baselines. On the Phi-2 model, mitigating the attribution errors by applying the corrective term allows us to safely deactivate $>$60\% of neurons in the model, with $<$3\% model accuracy loss. Furthermore, even if we further deactivate 80\% of neurons, we can still achieve a decently high model accuracy at 92\%. In contrast, most baselines of attribution metrics exhibit significant accuracy loss even when AR=80\%, due to the large impact of attribution errors as described in Section 2.3. Surprisingly, simply using the neurons' gradients as the attribution metric results in the lowest model accuracy with all ARs, due to the ignorance of neurons' output magnitudes.

In addition, we also notice that the model accuracy of some baseline schemes (e.g., IG and uncorrected GxO), exhibited slight drop when AR increases from 50\% to 80\%. This is because low sparsity reduces the learned noise and makes the model focusing more on the important knowledge [\citenum{hoefler2021sparsity}].

%

%
%

\begin{figure}
	\centering
	\begin{minipage}[b]{0.3\textwidth}
		\centering
		"You"
	\end{minipage}
	\hfill
	\begin{minipage}[b]{0.3\textwidth}
		\centering
		"will"
	\end{minipage}
	\hfill
	\begin{minipage}[b]{0.3\textwidth}
		\centering
		"grow"
	\end{minipage}
	\vfill
	\begin{minipage}[b]{0.3\textwidth}
		\centering
		\includegraphics[width=\linewidth, height=2cm]{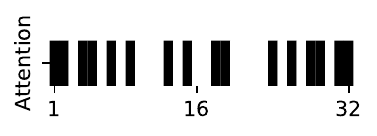}
	\end{minipage}
	\hfill
	\begin{minipage}[b]{0.3\textwidth}
		\centering
		\includegraphics[width=\linewidth, height=2cm]{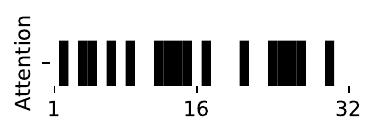}
	\end{minipage}
	\hfill
	\begin{minipage}[b]{0.3\textwidth}
		\centering
		\includegraphics[width=\linewidth, height=2cm]{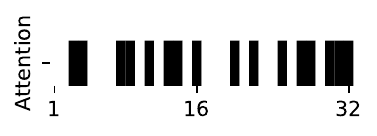}
	\end{minipage}
	\vfill
	\begin{minipage}[b]{0.3\textwidth}
		\centering
		\includegraphics[width=\linewidth, height=2cm]{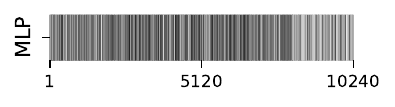}
	\end{minipage}
	\hfill
	\begin{minipage}[b]{0.3\textwidth}
		\centering
		\includegraphics[width=\linewidth, height=2cm]{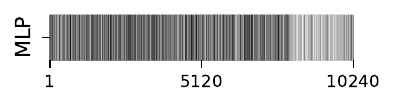}
	\end{minipage}
	\hfill
	\begin{minipage}[b]{0.3\textwidth}
		\centering
		\includegraphics[width=\linewidth, height=2cm]{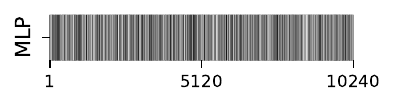}
	\end{minipage}
\vspace{-0.1in}
	\caption{The deactivation map of attention heads and MLP neurons over different input tokens on the Phi-2 model. AR=50\% and blocks in black indicate deactivated neurons.}
	\vspace{-0.2in}
	\label{fig:mask_different_sample}
\end{figure}

\vspace{-0.05in}
\subsection{Ablation Study}
\vspace{-0.05in}

To further investigate detailed characteristics of sparse activation in different model structures in SLMs, we conducted ablation studies on the attention layers and MLP layers in the Phi-2 model, respectively. Figure \ref{fig:mask_different_sample} visualizes the deactivated neurons in attention layers and MLP layers when generating different tokens from the Phi-2 model, and shows that different words and tokens within a sentence activate different attention heads and MLP neurons. Such diversity of neuron activation on different tokens justifies the benefits of sparse activation at run-time, which can adaptively adjust the activated neurons based on the current input sample, to achieve the best accuracy-sparsity tradeoff.

We noted that the number of MLP neurons in SLMs significantly exceeds the number of attention heads, and the characteristics of sparse activation in attention layers and MLP layers may hence be different. We further apply sparse activation only on attention layers and MLP layers, and the results with different activation ratios in Table \ref{tab:ablation_mlp} and Table \ref{tab:ablation_attention} show that MLP layers in SLMs are generally more over-parameterized than attention layers. While we can deactivate 80\% neurons in MLP layers with $<$3\% model accuracy loss, even deactivating 60\% neurons in attention layers will reduce the model accuracy to $<$90\%. This phenomenon is consistent with that reported in the existing work for LLMs [\citenum{liu2023deja}] and concurs that the impact of inter-layer dependency is much more in MLP layers than in attention layers, as we have shown in Figure \ref{fig: Average_inter_layer_dependency_ratio}.


\begin{table}[htbp]
	{\fontsize{8}{10}\selectfont
		\hspace{-0.1in}
		\begin{tabular}{crrrrrrrr}
			\toprule
			\textbf{Metric} & \textbf{AR=10\%} & \textbf{AR=20\%} & \textbf{AR=30\%} & \textbf{AR=40\%} & \textbf{AR=50\%} & \textbf{AR=60\%} & \textbf{AR=80\%} & \textbf{AR=100\%}\\
			\midrule
			Magnitude & 37.2 & 51.2 & 55.1 & 60.2 & 66.1 & 73.2 & 90.3 & 98.0 \\			
			IG & 71.2 & 75.8 & 78.7 & 78.0 & 77.9 & 75.0 & 73.2 & 98.0 \\
			GxO & 74.2 & 77.6 & 78.3 & 78.3 & 77.2 & 75.0 & 74.0 & 98.0 \\
			Corrected GxO & \textbf{\underline{91.4}} & \textbf{\underline{95.3}} & \textbf{\underline{96.2}} & \textbf{\underline{96.0}} & \textbf{\underline{93.6}} & \textbf{\underline{94.6}} & \textbf{\underline{98.0}} & 98.0 \\
			\bottomrule
	\end{tabular}}
	\vspace{0.05in}
	\caption{Accuracy of sparsely activated SLMs with neuron deactivation only in MLP layers. The Phi-2 model on the TruthfulQA dataset is used.}
	\vspace{-0.25in}
	\label{tab:ablation_mlp}
\end{table}

\begin{table}[htbp]
	{\fontsize{8}{10}\selectfont
		\hspace{-0.1in}
		\begin{tabular}{crrrrrrrr}
			\toprule
			\textbf{Metric} & \textbf{AR=10\%} & \textbf{AR=20\%} & \textbf{AR=30\%} & \textbf{AR=40\%} & \textbf{AR=50\%} & \textbf{AR=60\%} & \textbf{AR=80\%} & \textbf{AR=100\%}\\
			\midrule
			Magnitude & 58.8 & 67.0 & 75.6 & 78.2 & 83.4 & 85.5 & 92.2 & 98.0 \\			
			IG & 65.0 & 91.1 & 95.2 & 96.2 & 95.8 & 96.9 & 97.7 & 98.0 \\
			GxO & 64.4 & 90.8 & 96.1 & 96.1 & 97.0 & 97.0 & 97.7 & 98.0 \\
			Corrected GxO & \textbf{\underline{65.6}} & \textbf{\underline{79.9}} & \textbf{\underline{88.1}} & \textbf{\underline{89.9}} & \textbf{\underline{91.3}} & \textbf{\underline{91.4}} & \textbf{\underline{94.6}} & 98.0 \\
			\bottomrule
	\end{tabular}}
	\vspace{0.05in}
	\caption{Accuracy of sparsely activated SLMs with neuron deactivation only in attention layers. The Phi-2 model on the TruthfulQA dataset is used.} 
	\vspace{-0.15in}
	\label{tab:ablation_attention}
\end{table}

\vspace{-0.05in}
\subsection{Accuracy of Different SLMs with Sparse Activation}
\vspace{-0.05in}

We verify the generalizability of our proposed attribution metric over different LLMs, namely the models of both Phi and MobiLlama series. Results in Table \ref{tab:different_models} show that our corrective term, when being applied to the GxO metric, generally achieves much higher model accuracy on all the SLMs. In particular, we noticed that different SLMs are over-parameterized in different ways and hence exhibit highly different performances with the same sparse activation ratio. For example, even when 80\% of neurons are deactivated, Phi-1.5 exhibits the similar characteristics as Phi-2 did in Section 4.1 and only experiences 2\% accuracy loss. In contrast, MobiLlama models are much less over-parameterized, and have 10\% accuracy loss even when deactivating 20\% neurons. Nevertheless, our corrected GxO metric can still outperforms the best baseline by at least 25\% on the MobiLlama models.

\begin{table}
	{\fontsize{8}{10}\selectfont
		\hspace{-0.2in}
		\begin{tabular}{crrrrrrrr}					
			\toprule
			\makecell{\textbf{Model \&} \\ \textbf{Metric}} & \textbf{AR=10\%} & \textbf{AR=20\%} & \textbf{AR=30\%} & \textbf{AR=40\%} & \textbf{AR=50\%} & \textbf{AR=60\%} & \textbf{AR=80\%} & \textbf{AR=100\%} \\
			\midrule
			\rowcolor{gray!25}
			\tabularCenterstack{c}{\textbf{Phi-1.5}} \\
			Magnitude & 17.2 & 36.4 & 52.6 & 60.6 & 69.2 & 79.6 & 90.1 & 98.0 \\			
			IG & 80.8 & 89.9 & 92.5 & 93.7 & 93.6 & 93.5 & 92.4 & 98.0 \\
			GxO & 82.2 & 88.9 & 91.0 & 93.9 & 83.6 & 92.7 & 89.8 & 98.0 \\
			Corrected GxO & \textbf{\underline{79.6}} & \textbf{\underline{96.1}} & \textbf{\underline{97.0}} & \textbf{\underline{97.4}} & \textbf{\underline{97.6}} & \textbf{\underline{98.0}} & \textbf{\underline{97.7}} & 98.0 \\
			\rowcolor{gray!25}
		\midrule
			\tabularCenterstack{c}{\textbf{MobiLlama-0.5B}} \\
			Magnitude & 0.6 & 3.1 & 6.2 & 10.6 & 19.5 & 25.7 & 51.8 & 93.0 \\			
			IG & 4.7 & 7.2 & 11.1 & 10.9 & 7.4 & 4.7 & 5.7 & 93.0 \\
			GxO & 2.7 & 8.6 & 8.6 & 9.0 & 6.5 & 5.2 & 5.1 & 93.0 \\
				Corrected GxO & \textbf{\underline{3.2}} & \textbf{\underline{18.6}} & \textbf{\underline{34.8}} & \textbf{\underline{54.4}} & \textbf{\underline{59.4}} & \textbf{\underline{67.4}} & \textbf{\underline{83.0}} & 93.0 \\
			\rowcolor{gray!25}
			\midrule			
			\tabularCenterstack{c}{\textbf{MobiLlama-1B}} \\
			Magnitude & 6.4 & 9.8 & 17.4 & 20.2 & 23.9 & 29.8 & 57.4 & 91.3 \\			
			IG & 22.6 & 34.7 & 34.0 & 34.3 & 33.0 & 35.9 & 45.6 & 91.3 \\
			GxO & 20.1 & 26.4 & 31.1 & 30.0 & 30.3 & 31.6 & 34.6 & 91.3 \\
				Corrected GxO & \textbf{\underline{24.9}} & \textbf{\underline{49.0}} & \textbf{\underline{51.0}} & \textbf{\underline{62.7}} & \textbf{\underline{67.1}} & \textbf{\underline{69.1}} & \textbf{\underline{80.1}} & 91.3 \\
			\bottomrule
	\end{tabular}}
	\vspace{0.05in}
	\captionof{table}{Accuracy of sparsely activated SLMs with different activation ratios (AR). Different SLMs are evaluated on the TruthfulQA dataset.} 
	\vspace{-0.2in}
	\label{tab:different_models}
\end{table}

Such performance difference between Phi models and MobiLlama models is mainly because of their different model structures. The impact of inter-layer dependency is much less significant in Phi models, due to their have parallel transformer block structure (attention layer and MLP layer in a transformer block is parallel) [\citenum{gunasekar2023textbooks}]. In contrast, the MobiLlama models have sequential transformer block structure [\citenum{thawakar2024mobillama}] that leads to higher inter-layer dependency. This also justifies the effectiveness of our corrective term that aims to mitigate the impact of inter-layer dependency.

\begin{table}[H]
	\vspace{-0.1in}
	{\fontsize{8}{10}\selectfont
		\hspace{-0.1in}
		\begin{tabular}{crrrrrrrr}
			\toprule
			\makecell{\textbf{Dataset \&} \\ \textbf{Metric}} & \textbf{AR=10\%} & \textbf{AR=20\%} & \textbf{AR=30\%} & \textbf{AR=40\%} & \textbf{AR=50\%} & \textbf{AR=60\%} & \textbf{AR=80\%} & \textbf{AR=100\%} \\
			\midrule
			\rowcolor{gray!25}
			\vspace{0.05in}
			\textbf{TruthfulQA} \\
			Magnitude & 24.6 & 43.9 & 50.1 & 58.4 & 62.2 & 67.3 & 84.6 & 98.0 \\		
			IG & 56 & 66.8 & 72.4 & 74.6 & 74.5 & 72.0 & 70.7 & 98.0 \\
			GxO & 58.9 & 69.0 & 71.9 & 74.9 & 71.8 & 70.0 & 72.5 & 98.0 \\
			Corrected GxO & \textbf{\underline{85.5}} & \textbf{\underline{92.0}} & \textbf{\underline{93.6}} & \textbf{\underline{94.7}} & \textbf{\underline{95.1}} & \textbf{\underline{96.3}} & \textbf{\underline{98.0}} & 98.0 \\
			\rowcolor{gray!25}
			\midrule		
			\tabularCenterstack{c}{\textbf{Yahoo} \\ \textbf{Answers QA}} \\
			Magnitude & 22.0 & 36.2 & 48.0 & 55.0 & 63.8 & 69.7 & 90.0 & 100.0 \\			
			IG & 57.1 & 67.5 & 73.4 & 74.6 & 73.6 & 71.9 & 75.0 & 100.0 \\
			GxO & 50.0 & 67.8 & 69.4 & 74.2 & 70.9 & 66.5 & 70.3 & 100.0 \\
			Corrected GxO & \textbf{\underline{88.7}} & \textbf{\underline{93.6}} & \textbf{\underline{96.6}} & \textbf{\underline{99.2}} & \textbf{\underline{99.7}} & \textbf{\underline{99.2}} & \textbf{\underline{99.7}} & 100.0 \\
			\bottomrule
	\end{tabular}}
	\vspace{0.05in}
	\captionof{table}{Accuracy of Phi-2 model with different activation ratios (AR) on different datasets} 
	\vspace{-0.25in}
	\label{tab:different_datasets}
\end{table}

\vspace{-0.05in}
\subsection{Model Accuracy with Different Datasets}
\vspace{-0.05in}
We further evaluate the accuracy of sparsely activated SLMs with two different datasets, i.e., TruthfulQA and YahooAnswersQA, in very different knowledge domains. Results in Table \ref{tab:different_datasets} show that our approach outperforms other baseline methods on both datasets. The accuracy achieved by our corrected GxO metric on the YahooAnswersQA dataset are very similar to that on the Truthful QA dataset. This shows that our method is applicable to  various knowledge domains, including both fact checking and problem solving. Therefore, for SLMs focusing on QA tasks, our methods efficiently reduce the impact of inter-layer dependency and ensure efficient accuracy-sparsity tradeoffs. 

%

\vspace{-0.05in}
\subsection{Computational overhead of different attribution methods}
\vspace{-0.05in}

\begin{wrapfigure}{r}{0.3\linewidth}
	\vspace{-0.45in}
	\begin{minipage}[b]{\linewidth}
		\centering
		{\fontsize{8}{10}\selectfont
			\begin{tabular}{crrrrrrrr}
				\toprule
				\textbf{Method} & \textbf{time (s/token)}\\
				\midrule
				Magnitude & 0.11 \\                                    
				IG & 53.17\\
				GxO & 3.72  \\
				Corrected GxO & \textbf{\underline{3.72}} \\
				\bottomrule
		\end{tabular}}
		\vspace{0.05in}
		\captionof{table}{Computing overhead of different attribution metrics} 
		\vspace{-0.2in}
		\label{tab:ablation_attention}
	\end{minipage}
\end{wrapfigure}

Computing attribution scores could be expensive and incur extra computing costs. However, Table 7 show that the calculation of our proposed corrective term introduces negligible amounts of extra computing costs and hence retains the high compute efficiency of GxO attribution metric. In contrast, using Magnitude as the metric incurs the lowest overhead but results in large model accuracy loss.

\vspace{-0.1in}
\section{Conclusion}
\vspace{-0.1in}
In this paper, we aim to achieve sparse activation in SLMs. We demonstrated that the existing magnitude-based sparse activation cannot be applied to SLMs, and using gradient-based attribution scores for sparse activation is a better choice. We developed analytical methods that quantify and mitigate the attribution errors caused by inter-layer dependency of neurons' attribution scores, by applying a corrective term onto the existing GxO attribution metric. Our approach can achieve 80\% sparsification ratio on SLMs with $<$5\% accuracy loss, comparable to that on LLMs.

\setcitestyle{numbers}
\bibliographystyle{abbrvnat}
\bibliography{ref}


\newpage
\appendix

\section{Proof of Lemma \ref{lemma:3.3}}
\label{appendix: C}
\begin{proof}
	
	To prove this lemma, we first make an assumption that when the attribution scores of neurons in $L_2$ are changed by neuron deactivation in $L_1$, the signs of such changes of all neurons in $L_2$ are the same. That is, for any two neurons $j_1$ and $j_2$ in $L_2$, when neuron $i$ in $L_1$ is deactivated, we have
	\begin{equation}
		(S(h, g_{j_1}(\mathbf{X})-S(h, g_{j_1}(\widetilde{\mathbf{X}})) \cdot (S(h, g_{j_2}(\mathbf{X}))-S(h, g_{j_2}(\widetilde{\mathbf{X}}))>0.
	\end{equation}
	
	We verify this assumption with experiments. As shown in Table \ref{tab:ablation_attention}, on the Phi-2 model with the TruthfulQA dataset, when different activation ratio applies, most of neurons' attribution scores decrease, indicating negative changes of neurons' attribution scores.  

\begin{table}[htbp]
	\centering
	{\fontsize{8}{10}\selectfont
		\begin{tabular}{crrrrrrrr}
			\toprule
			\textbf{} & \textbf{AR=10\%} & \textbf{AR=20\%} & \textbf{AR=30\%} & \textbf{AR=40\%} & \textbf{AR=50\%} & \textbf{AR=60\%} & \textbf{AR=80\%} & \textbf{AR=90\%}\\
			\midrule
			Neuron ratio & 95.1 & 94.7 & 94.5 & 94.2 & 94.0 & 94.3 & 94.7 & 94.8 \\
			\bottomrule
	\end{tabular}}
\vspace{0.05in}
	\caption{The ratio of neurons with decreased importance scores under different ratios.} 
	\label{tab:ablation_attention}
\end{table}

	Based on this assumption, we want to prove that the error quantified in Eq. (\ref{eq:quantification_error}) has upper and lower bounds. The formulation of the error is as following
	\begin{equation}
	\sum_{i \in I(g(\mathbf{X}))} \text{S}(h, g_i(\mathbf{\widetilde{X}})) - \sum_{i \in I(g_i(\widetilde{\mathbf{X}}))} \text{S}(h, g_i(\widetilde{\mathbf{X}})).
	\end{equation}
	
	\noindent The first term \(\sum_{i \in I(g(\mathbf{X}))} \text{S}(h, g_i(\mathbf{\widetilde{X}}))\) is the sum of the smallest $N_m$ neurons importance scores in $L_2$ when considering the neuron deactivation in $L_1$, the second term \(\sum_{i \in I(g_i(\widetilde{\mathbf{X}}))} \text{S}(h, g_i(\widetilde{\mathbf{X}}))\) is the sum of the smallest $N_m$ neurons importance scores in $L_2$ without considering the neuron deactivation in $L_1$. The error caused by the inter-layer dependency can be calculated as the difference between the two terms which is the additional model output change. We can obtain the upper bound of the error by scaling.

	\begin{align}
	0 \leq \sum_{i \in I(g(\mathbf{X}))} \text{S}(h, g_i(\widetilde{\mathbf{X}})) 
	&- \sum_{i \in I(g_i(\widetilde{\mathbf{X}}))} \text{S}(h, g_i(\widetilde{\mathbf{X}})) \nonumber \\
	&= \sum_{i \in I(g(\mathbf{X}))} \{\text{S}(h, g_i(\mathbf{X})) + [\text{S}(h, g_i(\widetilde{\mathbf{X}})) - \text{S}(h, g(\mathbf{X}))]\} \\
	&\quad - \sum_{i \in I(g_i(\widetilde{\mathbf{X}}))} \{\text{S}(h, g_i(\mathbf{X})) + [\text{S}(h, g_i(\widetilde{\mathbf{X}})) - \text{S}(h, g_i(\mathbf{X}))]\}\\
	&\leq \sum_{i \in I(g(\mathbf{X}))} [\text{S}(h, g_i(\widetilde{\mathbf{X}})) - \text{S}(h, g_i(\mathbf{X}))] \\
	&\quad - \sum_{i \in I(g_i(\widetilde{\mathbf{X}}))} [\text{S}(h, g_i(\widetilde{\mathbf{X}})) - \text{S}(h, g_i(\mathbf{X}))]\\
	&= \sum_{i \in I(g(\mathbf{X})) \setminus I(g_i(\widetilde{\mathbf{X}}))} [\text{S}(h, g_i(\widetilde{\mathbf{X}})) - \text{S}(h, g_i(\mathbf{X}))]\\
	&\quad - \sum_{i \in I(g_i(\widetilde{\mathbf{X}})) \setminus I(g(\mathbf{X}))} [\text{S}(h, g_i(\widetilde{\mathbf{X}})) - \text{S}(h, g_i(\mathbf{X}))]\\
	&\leq \left| \sum_{i \in I(g(\mathbf{X}))} [\text{S}(h, g_i(\widetilde{\mathbf{X}})) - \text{S}(h, g_i(\mathbf{X}))] \right. \\
	&\quad + \left. \sum_{i \in I(g_i(\widetilde{\mathbf{X}}))} [\text{S}(h, g_i(\widetilde{\mathbf{X}})) - \text{S}(h, g_i(\mathbf{X}))] \right|\\
	&= \left| \sum_{i \in I(g(\mathbf{X})) \triangle I(g_i(\widetilde{\mathbf{X}}))} \left[\text{S}(h, g_i(\widetilde{\mathbf{X}})) - \text{S}(h, g_i(\mathbf{X}))\right] \right|\\
	&\leq \left| \sum_{i=1}^{N} [\text{S}(h, g_i(\widetilde{\mathbf{X}})) - \text{S}(h, g_i(\mathbf{X}))] \right|\\
	&= \left| \text{S}(h, g(\mathbf{X})) - \text{S}(h, g(\widetilde{\mathbf{X}})) \right|\\
	&= \left| \text{S}(F, \mathbf{X}) - \text{S}(F, \widetilde{\mathbf{X}}) \right|
	\end{align}

\section{Proof of Lemma \ref{lem:3.1}} 
\label{appendix: A}
\begin{proof}
	
	\(\text{S}(h, g(\mathbf{X}))\) can be represented as
	
	\begin{align}
		\text{S}(h, g(\mathbf{X})) &= \frac{\partial h(g(\mathbf{X}))}{\partial g(\mathbf{X})} g(\mathbf{X})^T \\
		&=\frac{\partial h(g(\mathbf{X}))}{\partial g(\mathbf{X})} \frac{\partial g(\mathbf{X})}{\partial \mathbf{X}}\mathbf{X}^T\\
		&=\frac{\partial h(g(\mathbf{X}))}{\partial \mathbf{X}} \mathbf{X}^T\\
		&=\frac{\partial F(\mathbf{X})}{\partial \mathbf{X}} \mathbf{X}^T\\
		&=\text{S}(F, \mathbf{X})
	\end{align}
	
\end{proof}

\section{Proof of Theorem \ref{Theorem:3.4}}
	\label{appendix: D}
	\noindent Since the impact of the intra-layer dependency is minimal, we can assume the neuron gradients in $L_1$ is not changed before and after applying masking as \(\frac{\partial F(\mathbf{X})}{\partial \mathbf{X}} = \frac{\partial F(\widetilde{\mathbf{X}})}{\partial \widetilde{\mathbf{X}}}\). Therefore, we can get the upper bound and lower bound by scaling as
	
	\begin{align}
	0 \leq \left|\text{S}(F, \mathbf{X}) - \text{S}(F, \mathbf{\widetilde{X}})\right| &= \left|(\frac{\partial F(\mathbf{X})}{\partial \mathbf{X}} \mathbf{X}^T)-(\frac{\partial F(\widetilde{\mathbf{X}})}{\partial \widetilde{\mathbf{X}}} \widetilde{\mathbf{X}}^T)\right|\\
	&= \left|\frac{\partial F(\mathbf{X})}{\partial \mathbf{X}}(\mathbf{X}^T - \widetilde{\mathbf{X}}^T)\right|\\
	&\leq ||\frac{\partial F(\mathbf{X})}{\partial \mathbf{X}}||_2 \cdot ||\mathbf{X}^T - \widetilde{\mathbf{X}}^T||_2\\
	&= |x_i| \cdot \sqrt{\sum_{k=1}^N  (\frac{\partial F}{\partial x_k})^2}
	\end{align}
\end{proof}

\noindent Given the definition of the error caused by inter-layer dependency, such error is no less than 0. Equality holds when the rank of neurons' attribution scores in $L_2$ is not changed when neuron deactivation in $L_1$, i.e., \(I(g(\mathbf{X})) = I(g(\widetilde{\mathbf{X}}))\). Therefore, the correction term is \( |x_i| \cdot \sqrt{\sum_{k=1}^N  (\frac{\partial F}{\partial x_k})^2}\), when the inter-layer dependency misleads to mask all neurons with positive importance score change but left the neurons with zero importance score change, while the neurons with zero importance score change are supposed to be masked without inter-layer dependency. Therefore, we conclude that lower bound of the error is 0 and the upper bound is $|x_i| \cdot \sqrt{\sum_{k=1}^N  (\frac{\partial F}{\partial x_k})^2}$.

\section{Details of Evaluation and Implementation Setup} 
\label{appendix: F}

To evaluate our method's performance in SLM, we focus on QA tasks, which are common and relevant for SLMs. Instead of multi-choice tasks, we use the task of sentence generation to provide a scenario for sparse activation on per token generation. We employ the TruthfulQA and YahooAnswersQA datasets, designed for sentence generation. The model input is a prompt, and the output is a sentence rather than a single word. We use BLEU as the evaluation metric because it is widely used for language model assessment and favors longer sentences, i.e., a short output of one or two words results in a very low BLEU score.

To ensure fair comparisons, we first generate outputs from the fully activated model for each SLM and then use these outputs as the ground truth when evaluating the sparsely activated model. This approach ensures that each fully activated model achieves nearly 100\% testing accuracy. 

We use the Phi and MobiLlama series models for comparison, as their similar sizes but different structures highlight our method's advantages. For instance, Phi-1.5 (1.3B) and MobiLlama-1B (1.2B) are comparably sized, yet Phi-1.5 employs a parallel transformer block, resulting in better performance.

Additionally, we apply sparse activation based on a uniform activation ratio. This means that for every attention layer and MLP layer, we deactivate the same percentage of attention heads or MLP neurons. The detailed rationale is explained in Section 3.3.

For the application of the proposed metric, we activate the attention heads or neurons whose outputs exceed a certain threshold. This threshold is determined by ranking the attribution scores of neurons. Since we apply the same deactivation ratio to each layer, the threshold varies between layers.


\end{document}